\documentclass[10pt,twocolumn,letterpaper]{article}

\usepackage{wacv}              

\usepackage{graphicx}
\usepackage{amsmath}
\usepackage{amssymb}
\usepackage{booktabs}

\usepackage{indentfirst}
\usepackage{times}
\usepackage{epsfig}
\usepackage{multirow}
\usepackage{color}
\usepackage[dvipsnames]{xcolor}
\usepackage{pifont}
\usepackage{afterpage}
\usepackage{xcolor}
\usepackage{pifont}
\usepackage{bbding}
\newcommand{\cmark}{\ding{51}} 

\usepackage{array}
\newcolumntype{L}[1]{>{\raggedright\let\newline\\\arraybackslash\hspace{0pt}}m{#1}}
\newcolumntype{C}[1]{>{\centering\let\newline\\\arraybackslash\hspace{0pt}}m{#1}}
\newcolumntype{R}[1]{>{\raggedleft\let\newline\\\arraybackslash\hspace{0pt}}m{#1}}

\definecolor{hydra_attention_color}{RGB}{0,154,85}
\definecolor{ubin_red}{rgb}{0.8, 0, 0}

\usepackage[pagebackref,breaklinks,colorlinks]{hyperref}

\usepackage[capitalize]{cleveref}
\crefname{section}{Sec.}{Secs.}
\Crefname{section}{Section}{Sections}
\Crefname{table}{Table}{Tables}
\crefname{table}{Tab.}{Tabs.}

\def\wacvPaperID{370} 
\def\confName{WACV}
\def\confYear{2024}

\begin{document}

\title{MetaSeg: MetaFormer-based Global Contexts-aware Network for Efficient Semantic Segmentation}

\author{Beoungwoo Kang\thanks{These authors contributed equally.}   , Seunghun Moon\footnotemark[1]   , Yubin Cho\footnotemark[1]   , Hyunwoo Yu\footnotemark[1]   , and Suk-Ju Kang\\
Sogang University, Republic of Korea\\
{\tt\small \{beoungwoo, moonsh97, dbqls1219, hyunwoo137, sjkang\}@sogang.ac.kr}
}
\maketitle

\begin{abstract}
   Beyond the Transformer, it is important to explore how to exploit the capacity of the MetaFormer, an architecture that is fundamental to the performance improvements of the Transformer.
   Previous studies have exploited it only for the backbone network. Unlike previous studies, we explore the capacity of the Metaformer architecture more extensively in the semantic segmentation task.
   We propose a powerful semantic segmentation network, MetaSeg, which leverages the Metaformer architecture from the backbone to the decoder. Our MetaSeg shows that the MetaFormer architecture plays a significant role in capturing the useful contexts for the decoder as well as for the backbone. In addition, recent segmentation methods have shown that using a CNN-based backbone for extracting the spatial information and a decoder for extracting the global information is more effective than using a transformer-based backbone with a CNN-based decoder. This motivates us to adopt the CNN-based backbone using the MetaFormer block and design our MetaFormer-based decoder, which consists of a novel self-attention module to capture the global contexts. To consider both the global contexts extraction and the computational efficiency of the self-attention for semantic segmentation, we propose a Channel Reduction Attention (CRA) module that reduces the channel dimension of the query and key into the one dimension. In this way, our proposed MetaSeg outperforms the previous state-of-the-art methods with more efficient computational costs on popular semantic segmentation and a medical image segmentation benchmark, including ADE20K, Cityscapes, COCO-stuff, and Synapse. The code is available at \url{https://github.com/hyunwoo137/MetaSeg}.
\end{abstract}

\section{Introduction}
\label{sec:intro}

Semantic segmentation, which classifies categories for each pixel, is a challenging task in computer vision. This task has a wide range of applications \cite{yu2022vision, cho2022class, cho2023cross}, including autonomous driving and medical image segmentation.

With the great success of the vision transformer (ViT) \cite{dosovitskiy2020image} in the image classification, the transformer-based methods have been introduced in the field of semantic segmentation. 
Most previous studies \cite{wang2021pyramid, wu2021cvt, xie2021segformer, wang2022pvt, yu2024embedding} mainly utilize the self-attention layer in the transformer block to achieve the superior performance. However, recent research \cite{yu2022metaformer} found that the abstracted architecture of the transformer block (\textit{i.e.}, MetaFormer block), which consists of a token-mixer, channel MLPs and residual connections, plays a more significant role in achieving the competitive performance than the specific token mixer (\textit{e.g.} attention, spatial MLP). 
Therefore, the MetaFormer architecture has the potential to be variably applied with different token mixers depending on the specific purpose.

\begin{figure}[t]
\includegraphics[width=0.98\linewidth]{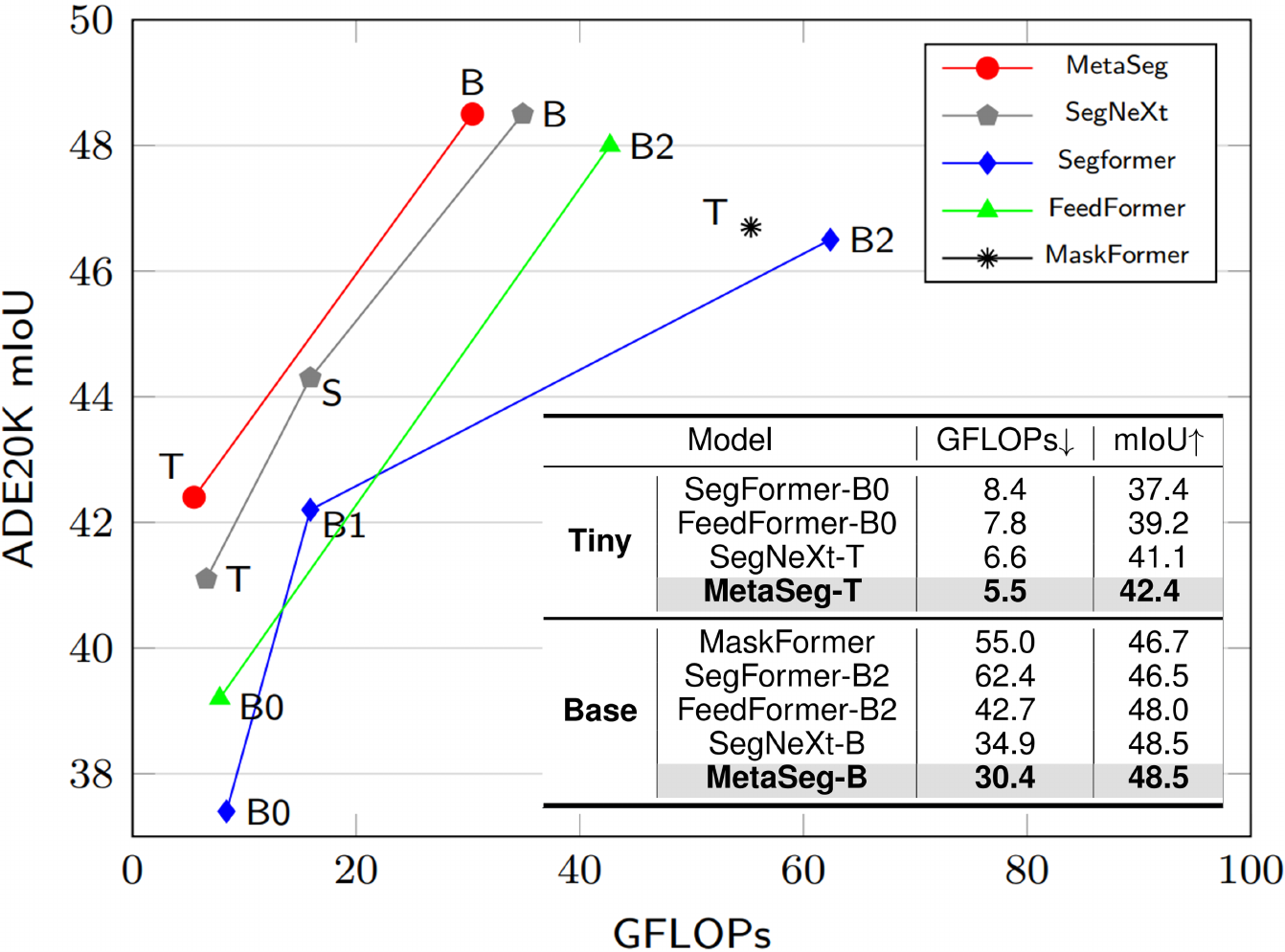} 
\caption{Performance-Computation curves on ADE20K validation set. Compared the performance and computation of our MetaSeg with recent models \cite{guo2022segnext,xie2021segformer, shim2023feedformer, cheng2021per}. We find that our MetaSeg has the best trade-off between the performance and computational costs.}
\label{sim_intro}
\end{figure}

 From the MetaFormer architecture, some recent studies \cite{li2022efficientformer, wang2023internimage} have derived their own methods. For example, EfficientFormer \cite{li2022efficientformer} employs the MetaFormer architecture using the self-attention as the token mixer to effectively capture the global semantic-aware features. InternImage \cite{wang2023internimage} also utilizes the MetaFormer with the deformable convolution as the token mixer to capture the contextual information. These methods have exploited the capacity of the MetaFormer architecture only for the encoder. 
 However, unlike previous studies, we take advantage of the capacity of the MetaFormer block more extensively for the semantic segmentation task. Therefore, we propose a novel and powerful segmentation network, \textit{MetaSeg}, which utilizes the MetaFormer block up to the decoder to obtain the enhanced visual representation.

In addition, previous segmentation methods \cite{wang2021pyramid, xie2021segformer, liu2021swin} used the transformer-based backbone with the CNN-based decoder. However, recent studies \cite{guo2022segnext, zhang2022topformer} have shown that using the CNN-based backbone for extracting the local information and the decoder for extracting the global information is more effective in improving the performance by compensating for the globality in the local contexts. Based on this observation, we adopt the CNN-based backbone (\textit{i.e.}, MSCAN \cite{guo2022segnext}) that contains the MetaFormer block used the convolution as a token mixer, and design a novel transformer-based decoder. Since it is important to consider the globality in the decoder to complement the CNN-based encoder features, the proposed decoder leverages the MetaFormer block that uses the self-attention as a token mixer to capture the global contexts. However, the self-attention has a limitation of the considerable computational costs due to the high-resolution features in the semantic segmentation task.

To address this issue, we propose a novel and efficient self-attention module, \textit{Channel Reduction Attention }(CRA), which embeds the channel dimension of the query and key into the one dimension for each head in the self-attention operation. 
Conventional self-attention methods \cite{dosovitskiy2020image, wang2021pyramid, liu2021swin, wu2021cvt, xie2021segformer}, which embed the channel dimension of the query and key without the channel reduction, show great performance but have high computational costs. 
Compared to these methods, our method leads to competitive performance with the computational reduction. This indicates that our CRA can sufficiently consider the globality even when each query and key token is a scalar type, not a vector. 
Therefore, our CRA module is more efficient and effective than the previous self-attention modules.

To demonstrate the effectiveness and efficiency of our method, we conduct experiments on the challenging semantic segmentation datasets: ADE20K \cite{zhou2017scene}, Cityscapes \cite{cordts2016cityscapes}, and COCO-stuff \cite{caesar2018coco}. To verify the ability for the application, we also conduct experiments on the medical image segmentation dataset: Synapse \cite{landman2015synapse}. As shown in Fig. \ref{sim_intro}, our MetaSeg-T and MetaSeg-B surpass the previous state-of-the-art methods on three public semantic segmentation benchmarks, including ADE20K, Cityscapes, and COCO-Stuff. Especially, our MetaSeg-T outperforms SegNeXt-T \cite{guo2022segnext} by 1.3\%, 0.3\% and 1.0\% mIoU improvements with 16.7\%, 5.2\% and 16.7\% lower computational costs on ADE20K, Cityscapes, and COCO-Stuff, respectively.

In summary, the main contributions of our method are summarized as follows.

\begin{itemize}
    \item The proposed MetaSeg is a powerful semantic segmentation network that effectively captures the local to global contexts, showing that the capacity of the MetaFormer architecture can be extended to the decoder as well as the encoder.
    \item We propose Channel Reduction Attention (CRA), a novel and efficient self-attention module for semantic segmentation, which can consider the globality efficiently by reducing the channel dimension of the query and key into the one dimension for the computational reduction in the self-attention operation.  
    \item Our proposed MetaSeg outperforms the previous state-of-the-art methods in terms of efficiency, accuracy and robustness on three challenging semantic segmentation datasets and a medical image segmentation dataset to show ours applicability across different domains.
\end{itemize}

\section{Related Works}
\subsection{MetaFormer-based architecture}

MetaFormer is an general architecture of the transformer \cite{vaswani2017attention} where the token mixer is not specified. Recent methods \cite{tolstikhin2021mlp, touvron2022resmlp, yu2022metaformer} have explored various types of token mixers within the MetaFormer architecture to encourage the performance. Mlp-Mixer \cite{tolstikhin2021mlp} and ResMLP \cite{touvron2022resmlp} utilized MLP-like token mixers. PoolFormer \cite{yu2022metaformer} simply exploited pooling as token mixers to verify the power of the MetaFormer architecture. PVT \cite{wang2021pyramid}, Swin \cite{liu2021swin}, CvT \cite{wu2021cvt}, and EfficientFormer \cite{li2022efficientformer} adopted the self-attention as token mixers to aggregate the global information. These studies have focused on exploiting a variant token mixer based on the MetaFormer in the encoder. Therefore, we propose novel MetaFormer block which is leverage our Channel Reduction Attention (CRA) module as a token mixer. In addition, unlike the previous methods that apply the MetaFormer architecture to the encoder, we propose novel approach that the capacity of the MetaFormer architecture is extended to the decoder to consider the globality that is helpful for improving the segmentation performance.

\begin{figure*}[t]
\includegraphics[width=0.97\linewidth]{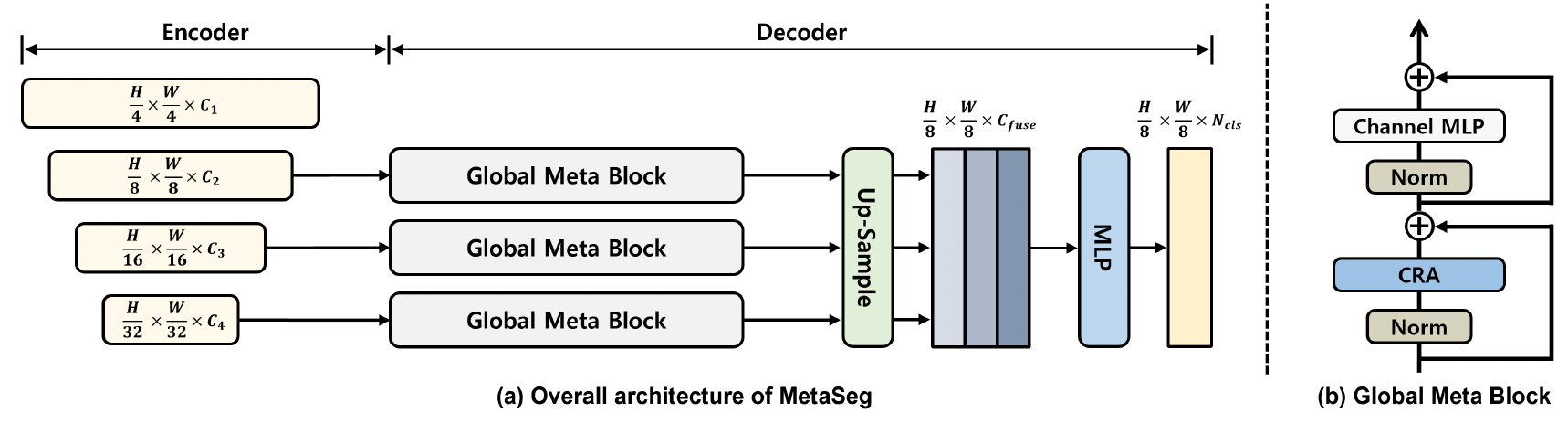} 
\caption{(a) Overall architecture of MetaSeg, consisting of two main part: hierarchical CNN-based Encoder and Global Meta Blcok (GMB) based decoder. (b) Details of the GMB, which is composed with the proposed Channel Reduction Attention (CRA) module and the channel MLP. Our MetaSeg extracts the multi-scale feature that contains local information in the encoder and complements the global information in the GMB of the decoder.}
\label{model_1}
\end{figure*}

\subsection{Semantic segmentation}
As ViT \cite{dosovitskiy2020image} have achieved the great success on the image classification task, self-attention based transformer backbones have also been explored in the semantic segmentation task. SETR \cite{zheng2021rethinking} was the first to use ViT as a backbone on the segmentation task. PVT \cite{wang2021pyramid}, Swin \cite{liu2021swin}, CvT \cite{wu2021cvt}, and LeViT \cite{graham2021levit} studied the hierarchical transformer-based backbone to exploit the multi-scale features. Beyond introducing transformer backbones for the segmentation, Segformer \cite{xie2021segformer} designed a light-weight transformer backbone and a MLP-based decoder to consider the computational efficiency. More recent methods \cite{zhang2022topformer, guo2022segnext} adopted the CNN-based backbone with the transformer-based decoder to aggregate the local to global information. TopFormer \cite{zhang2022topformer} encoded the tokens by the MobileNetV2 \cite{sandler2018mobilenetv2}, and then fed the tokens into the transformer blocks. In SegNeXt \cite{guo2022segnext}, the convolution-based encoder extracts the spatial information and the transformer-based decoder extracts the global context. These methods \cite{guo2022segnext, zhang2022topformer} have demonstrated that using the CNN-based backbone with the transformer-based decoder is effective for the semantic segmentation. According to these studies, we adopt the combination of the CNN-based backbone and transformer-based decoder. 

 Additionally, transformer-based segmentation methods \cite{liu2021swin, wang2021pyramid, xie2021segformer} have considered the computational efficiency of the attention mechanism due to high-resolution features. Swin \cite{liu2021swin} proposed a shifted window self-attention by partitioning the feature maps into the windows. Some recent methods \cite{wang2021pyramid,xie2021segformer} adopted a spatial reduction attention that reduces the resolution of the key-value. In this paper, we introduce a novel self-attention module, Channel Reduction Attention (CRA), which reduces the channel dimension of the query and key into the one dimension for efficient computational costs of the self-attention.

\section{Method}
\label{sec:formatting}
This section describes our MetaSeg architecture, an efficient and powerful segmentation network. Basically, we adopt the CNN-based encoder and MetaFormer-based decoder to aggregate the local and global information. We first explain the overall architecture, and then explain the encoder and decoder. Finally, we describe the Global Meta Block (GMB) with the proposed Channel Reduction Attention (CRA) that is an efficient self-attention module.

\subsection{Overall Architecture}
As shown in Fig. \ref{model_1} (a), our MetaSeg is based on the MetaFormer block with a hierarchical backbone network of the four stages. We utilize the CNN-based encoder that adopts a series of convolutional layers as a token mixer. The encoder aggregates the local information from the input via the token mixer. For the decoder, we design the novel CRA module as a token mixer to capture the global contexts with low computational costs.

\subsubsection{Hierarchical convolutional encoder}
We adopt the CNN-based pyramid encoder to acquire multi-scale features. Following previous encoder-decoder structured segmentation networks, given an image $I \in \mathbb{R}^{H \times W \times 3}$ as an input, each stage of the encoder extracts the down-sampled features $F_i \in \mathbb{R}^{\frac{H}{2^{i+1}} \times \frac{W}{2^{i+1}} \times {C_i}}$  where $i \in \{1,2,3,4\}$ and $C_i$ denote the index of the encoder stage and the channel dimension. These features provide the coarse to fine-grained features that leads to the performance improvements of the semantic segmentation. Specifically, we adopt MSCAN \cite{guo2022segnext} as a encoder, which consists of MetaFormer blocks using a convolution-based token mixer. 

\subsubsection{Lightweight decoder}

\begin{figure*}[t]
\centering
\includegraphics[width=0.97\linewidth]{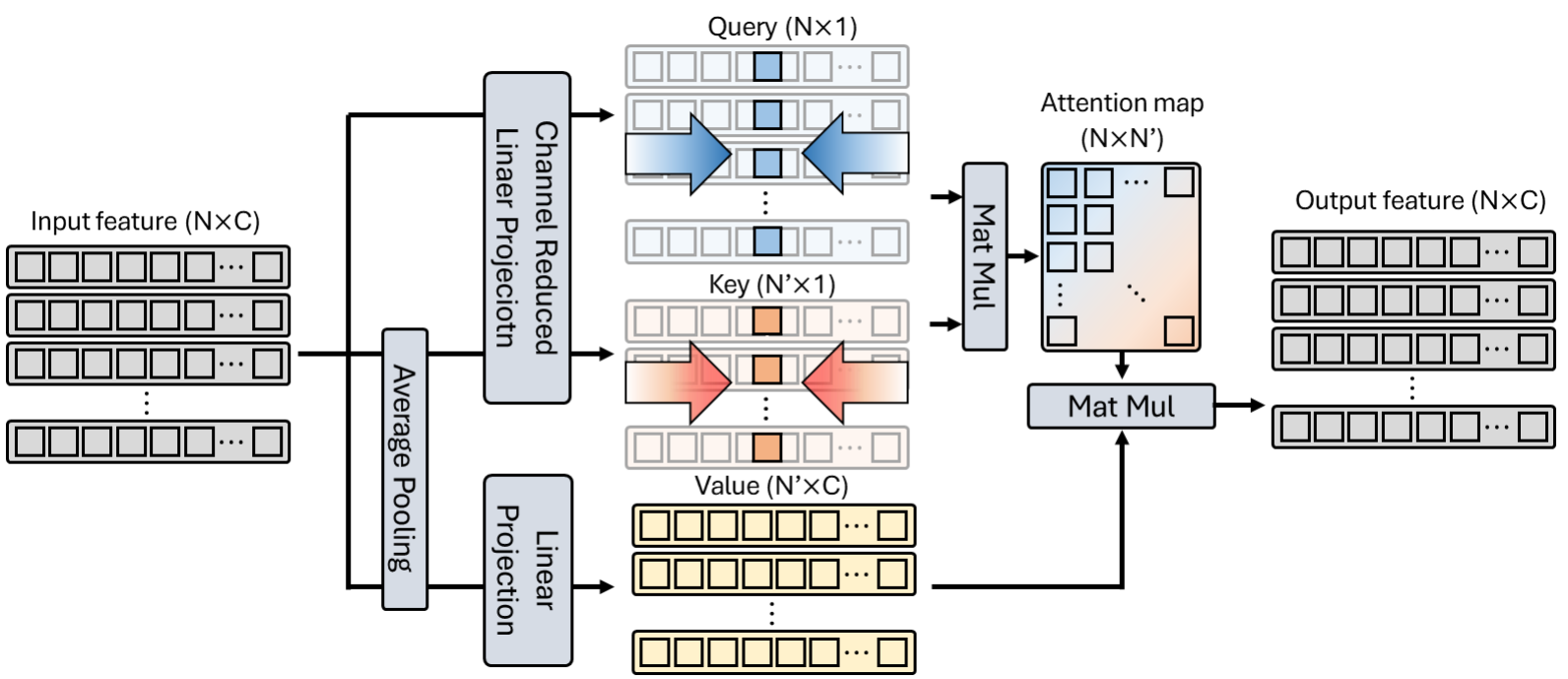} 
\caption{Illustration of the proposed Channel Reduction Attention (CRA). In our CRA, the channel dimension of the query and key is reduced to the one dimension for the computational efficiency and our CRA can capture the globality of the features effectively.}
\label{module_1}
\end{figure*}

The decoder of our MetaSeg exploits the MetaFormer architecture to improve the capture the global contexts that are not considered enough in the encoder. We discovered that the MetaFormer block, with the self-attention module as a token mixer, exhibits exceptional capability in gathering global contexts from the multi-scale features of the encoder. The decoder consists of following components: the Global Meta Block (GMB), the up-sampling layer, the MLP layer and the prediction layer. The up-sampling layer expands the feature resolution to $\frac{H}{8} \times \frac{W}{8}$, unifying the size of outputs extracted from the GMB of each stage. We exclude the features of the first encoder stage since they contain too much low-level information and bring high computational costs. The MLP layer then concatenates the up-sampled features. Finally, the prediction layer predicts the segmentation mask. The overall procedure in decoder is as follows:
\begin{equation}
\centering
\begin{split}
    &\hat{F}_i = \mathrm{GMB}(F_i),\ i \in \{2,3,4\} \\
    &F_{up\_i} = \mathrm{UpSample}(\frac{H}{8} \times \frac{W}{8})(\hat{F}_i),\ C_{fuse}=\sum_{i=2}^4 C_i\\
    &F = \mathrm{Linear}(C_{fuse}, C_{MLP})(\mathrm{Concat}(F_{up\_i})),  \\
    &Z = \mathrm{Linear}(C_{MLP}, N_{cls})(F), \\
\end{split}
\end{equation}
where $\mathrm{Linear}(a,b)(\cdot)$ denotes a linear layer with a size of $a$ as input dimensions and a size of $b$ as output dimensions. $C_{MLP}$ denotes the channel dimension of the MLP. $N_{cls}$ is defined as the number of classes.

\subsection{Global Meta Block (GMB)}
The proposed GMB leverages the MetaFormer block in the decoder to further enhance the global contexts of the feature representations extracted by the encoder, which mainly focuses on the local context. As illustrated in Fig. \ref{model_1} (b), the GMB adopts the MetaFormer block of two residual sub-blocks and employs a novel channel reduction self-attention (CRA) module as a token mixer. Our CRA module effectively captures global contexts of the features with efficient computational costs. The GMB is performed at each stage except the first stage (\textit{i.e.}, $i \in \{2,3,4\}$). The overall operation is defined as follows:
\begin{equation}
\centering
\begin{split}
    &M_i = \mathrm{CRA(LN}(F_i)) + F_i\ , \\
    &\hat{F}_i = \mathrm{MLP(LN}(M_i)) + M_i\ ,\\
\end{split}
\end{equation}
where $\mathrm{LN}$ and $\mathrm{MLP}$ denote the layer normalization and the channel MLP layer, respectively. 

\begin{table*}[t]\centering
\renewcommand{\arraystretch}{}
\resizebox{\textwidth}{!}{

\begin{tabular}{L{3.5cm}|c|ccc|ccc|ccc}
\toprule[1.5pt]

\multirow{2}{*}{{Method}}&\multirow{2}{*}{{Params(M)}}& \multicolumn{3}{c|}{ADE20K}& \multicolumn{3}{c|}{Cityscapes}& \multicolumn{3}{c}{COCO-Stuff} \\ & & {{GFLOPs $\downarrow$}} & \multicolumn{2}{c|}{mIoU (SS/MS) $\uparrow$}  & {{GFLOPs $\downarrow$}} & \multicolumn{2}{c|}{mIoU (SS/MS) $\uparrow$} & {{GFLOPs $\downarrow$}} & \multicolumn{2}{c}{mIoU (SS/MS) $\uparrow$} \\

\midrule
SegFormer-B0 \cite{xie2021segformer} &3.8 &8.4 &37.4 &38.0 &125.5 &76.2 &78.1 &8.4 &35.6 & - \\
FeedFormer-B0 \cite{shim2023feedformer} &4.5 &7.8 &39.2 & - &107.4 &77.9 & -  & -  & -  & -  \\
SegNeXt-T \cite{guo2022segnext}   &4.3 &6.6 &41.1 &42.2 &50.5  &79.8 &81.4 &6.6 &38.7 &39.1 \\
\midrule
\textbf{MetaSeg-T (Ours)}  &{4.7} &\textbf{5.5} &\textbf{42.4} &\textbf{43.4}  &\textbf{47.9}  &\textbf{80.1} &\textbf{81.5} &\textbf{5.5} &\textbf{39.7} &\textbf{40.2} \\

\midrule[0.05cm]
SegFormer-B2 \cite{xie2021segformer}  &27.5 &62.4 &46.5 &47.5 &717.1 &81.0 &82.2  &62.4 &44.6 & - \\
MaskFormer \cite{cheng2021per}    &42.0 &55.0 &46.7 &48.8  & - & - & - & -  & - & -  \\
FeedFormer-B2 \cite{shim2023feedformer} &29.1 &42.7 &48.0 & -  &522.7 &81.5 & - & -   & -  & - \\
SegNeXt-B \cite{guo2022segnext}     &27.6 &34.9 &48.5 &\textbf{49.9}  &275.7 &82.6 &83.8 &34.9   &45.8 &46.3 \\
\midrule
\textbf{MetaSeg-B (Ours)}     &{29.6} &\textbf{30.4} &\textbf{48.5} &{49.4} &\textbf{251.1} &\textbf{82.7} &\textbf{83.8}  &\textbf{30.4} &\textbf{45.8} &\textbf{46.3} \\

\bottomrule[1.5pt]
\end{tabular}
}
\caption{Comparison of our MetaSeg with previous state-of-the-arts methods on ADE20K, Cityscapes and COCO-Stuff. GFLOPs is calculated with $512 \times 512$ resolutions for ADE20K and COCO-Stuff, $2048 \times 1024$ resolutions for Cityscapes. Compared to previous state-of-the-arts methods, our MetaSeg model displays great effectiveness and efficiency.}
\label{table_1}
\end{table*}

\subsubsection{Channel Reduction Attention}
We propose the Channel Reduction Attention (CRA) module as a novel token mixer utilized in the GMB to consider both the globality extraction and the computational efficiency of the self-attention for the semantic segmentation.
Our CRA is based on the multi-head self-attention. The key and value are average pooled before the attention operation. As shown in Fig. \ref{module_1}, the channel dimensions of the query and key are embedded into the one dimension to further reduce the computational costs. We found that the channel squeezed query 
$Q \in \mathbb{R}^{\mathrm{Head} \times H_i W_i \times 1}$ and key $K \in \mathbb{R}^{{\mathrm{Head} \times ({H_i W_i}/{r_i^2}}) \times 1}$ can sufficiently extract global similarities.  The CRA operation is formulated as follows:
\begin{equation}
\centering
\begin{split}
    &\mathrm{CRA}(F_i) = \mathrm{Concat(Head_0, ... , Head_{j})}W^O_i\ ,\\
    &Q_i = F_iW_j^Q,\ K_i = \mathrm{AvgPool}(F_i)W_j^K \ ,\\
    &V_i = \mathrm{AvgPool}(F_i)W_j^V,\ \mathrm{Head_j} = \mathrm{Att}(Q_i, K_i, V_i)\ , \\ &\mathrm{Att}(Q_i,K_i,V_i) = \mathrm{Softmax}({Q_i K_i^T})V_i\ ,
\end{split}
\end{equation}
where $W_j^Q, W_j^K \in \mathbb{R}^{C_i \times 1}$, $W_j^V \in \mathbb{R}^{C_i \times \frac{C_i}{j}}$ and $W^O \in \mathbb{R}^{C_i \times C_i}$ are projection parameters. $j$ denotes the number of attention heads. $\mathrm{AvgPool}$ is the average pooling of scale $r_i\in\{2,4,8\}$ at each stage, respectively. Compared to SRA\cite{wang2021pyramid} that is a previous efficient self-attention method, the computational complexity of our CRA is as:
\begin{equation}
\centering
\begin{split}
    N'=\frac{N}{r_i}\ ,\ &\Omega(SRA) = (N')^2 C + (N')^2 C\ , \\
    &\Omega(CRA) = (N')^2 1 + (N')^2 C\ , \\
\end{split}
\end{equation}
where $N$ denotes the number of pixel tokens. In eq.(4), the left and right terms indicate the computations of the query-key operation and the computations of the attention weight-value operation, respectively. By reducing the computation of the query-key operation by C times, our CRA reduces the total computation of the attention operation by about twice.

%

\subsubsection{Channel MLP}
The channel MLP is used to consolidate the features processed with our token mixer. Channel MLP consists of the two 1$\times$1 convolution layers with a GELU activation layer. The operation is defined as follows:
\begin{equation}
\centering
\begin{split}
    &\mathrm{MLP}(x) = \mathrm{Conv_{1 \times 1}}(\mathrm{GELU(Conv_{1 \times 1}}(x)))\ , \\
\end{split}
\end{equation}
where $\mathrm{Conv}_{1 \times 1}$ denotes the $1 \times 1$ convolution layer.

\section{Experiment}
\subsection{Experimental Settings}
\noindent
\textbf{Datasets.}
We conducted experiments on four publicly available datasets, ADE20K \cite{zhou2017scene}, Cityscapes \cite{cordts2016cityscapes}, COCO-Stuff \cite{caesar2018coco}, and Synapse \cite{landman2015synapse}. ADE20K is a challenging scene parsing dataset composed of 20,210/2,000/3,352 images for training, validation, and testing with 150 semantic categories. Cityscapes is an urban driving scene dataset that contains 5,000 images finely annotated with 19 categories. It composed of 2,975/500/1,525 images in training, validation, and testing. COCO-Stuff is also a challenging dataset, which contains 172 semantic categories and 164,062 images. Synapse is an abdominal organ dataset that consists 30 Computerized Tomography (CT) scans with 3779 axial contrast-enhanced abdominal CT images. Following the experimental settings of TransUNet \cite{chen2021transunet}, we split the Synapse dataset into 18 scans for training, and 12 for validation.


\begin{table}[t]
\centering
\renewcommand{\arraystretch}{}
\resizebox{\columnwidth}{!}{
\begin{tabular}{l|c|c|c|c}
\toprule[1.4pt]
Method & Params (M) & {{GFLOPs $\downarrow$}} & {{mIoU (\%) $\uparrow$}} & {{FPS $\uparrow$}} \\
\midrule
SegFormer-B0 \cite{xie2021segformer} & 3.8  & 51.8 & 74.2 & 25.5  \\

FeedFormer-B0 \cite{shim2023feedformer}  & 4.5 & 41.6 \textcolor{hydra_attention_color}{(-19.7\%)}& 75.5 & 28.9 \textcolor{hydra_attention_color}{(+13.3\%)}\\

SegNeXt-T \cite{guo2022segnext}& 4.3 & 29.3 \textcolor{hydra_attention_color}{(-43.4\%)} & 77.8 & 30.2 \textcolor{hydra_attention_color}{(+18.4\%)} \\


\midrule
\textbf{MetaSeg-T (Ours)} & {4.7} & \textbf{26.2} \textcolor{hydra_attention_color}{\textbf{(-49.4\%)}} & \textbf{78.4} & \textbf{33.6} \textcolor{hydra_attention_color}{\textbf{(+31.8\%)}} \\
\bottomrule[1.4pt]
\end{tabular}
}
\caption{FPS comparison with recent state-of-the-art methods at the input size of 1536$\times$768 using a RTX3090 GPU on Cityscapes.} 
\label{table_6}
\end{table}


\noindent
\textbf{Implementation details.} 
The mmsegmentation codebase was used to train our model on 4 RTX 3090 GPUs. We used MSCAN \cite{guo2022segnext} as a backbone network. Our model with MSCAN-T and MSCAN-B backbones were each named MetaSeg-T, MetaSeg-B, and our decoder was randomly initialized. For semantic segmentation evaluation, we adopted the mean Intersection over Union (mIoU) for ADE20K, Cityscapes, and COCO-Stuff datasets, and the Dice Similarity Score (DSC) for Synapse dataset. During the training, we applied the commonly used data augmentation such as random horizontal flipping, random scaling from 0.5 to 2.0 ratios and random cropping with the size of 512$\times$512, 1024$\times$1024, and 512$\times$512 for ADE20K, Cityscapes, and COCO-Stuff datasets, respectively. For Synapse dataset, we used random rotation and flipping for data augmentation with the size of 224$\times$224. We trained our models using AdamW optimizer for 160K iterations on ADE20K and Cityscapes, 160K iterations on COCO-Stuff, and 30K iterations on Synapse. The batch size was 16 for ADE20K and COCO-Stuff, 8 for Cityscapes, and 24 for Synapse. The poly LR schedule with a factor of 1.0 and an initial learning rate of 6e-5 were used.

\begin{table}[t]
\centering
\renewcommand{\arraystretch}{}
\resizebox{\columnwidth}{!}{ 
\footnotesize

\begin{tabular}{L{4cm}|C{2.6cm}}
\toprule[1.pt]

Method & DSC (\%) $\uparrow$ \\
\midrule
V-Net \cite{milletari2016v}        &68.81 \\
DARR \cite{fu2020domain}         &69.77 \\
UNet \cite{ronneberger2015u}         &70.11 \\
R50+ViT \cite{dosovitskiy2020image}      &71.29 \\
AttnUNet \cite{schlemper2019attention}     &71.70 \\
R50+UNet \cite{chen2021transunet}     &74.68 \\
R50+AttnUNet \cite{chen2021transunet} &75.57 \\
TransUNet \cite{chen2021transunet}    &77.48 \\
MT-UNet \cite{wang2022mixed}      &78.59 \\
SwinUNet \cite{cao2022swin}     &79.13 \\
HiFormer \cite{heidari2023hiformer}     &80.69 \\
\midrule
\textbf{MetaSeg-B (Ours)} &\textbf{82.78} \\

\bottomrule[1.pt]
\end{tabular}
}
\caption{Comparison with the previous state-of-the-art methods on Synapse dataset.}
\label{table_2}
\end{table}

\subsection{Comparison with State-of-the-Art Methods}

\noindent
\textbf{ADE20K, Cityscapes, and COCO-Stuff datasets.}
In Table \ref{table_1}, we compared our MetaSeg performance with previous state-of-the-art methods on ADE20K, Cityscapes, and COCO-Stuff datasets. This comparison includes the number of the parameters, Floating Point Operations (FLOPs), and mIoU under both the single scale (SS) and multi-scale (MS) flip inference strategies. As shown in the Table \ref{table_1}, MetaSeg-T showed significant performance of 42.4\% mIoU with only 4.7M parameters and 5.5 GFLOPs for ADE20K. Compared to SegNeXt-T that uses the same backbone \cite{guo2022segnext}, our MetaSeg-T achieved 1.3\% higher mIoU and 16.7\% lower GFLOPs on ADE20K. Moreover, our MetaSeg-T showed 0.3\% and 1.0\% higher mIoU with 5.2\% and 16.7\% lower GFLOPs on Cityscapes and COCO-Stuff, respectively. Our larger model, MetaSeg-B, also achieved competitive performance compared to previous state-of-the-art models. MetaSeg-B showed 48.5\% mIoU with 12.9\% less computations compared to SegNeXt-B on ADE20K. Furthermore, our MetaSeg-B achieved 82.7\% and 45.8\% mIoU with 8.9\% and 12.9\% less GFLOPs on Cityscapes and COCO-Stuff, respectively. These results demonstrated that our MetaSeg effectively captures the local to global contexts by leveraging the MetaFormer architecture up to the decoder with an efficient token mixer, our CRA.

\begin{table}[t]
\centering
\renewcommand{\arraystretch}{}
\resizebox{\columnwidth}{!}{
\begin{tabular}{l|l|c|cc}
\toprule[1.4pt]

\multirow{2}{*}{{Backbone}}&\multirow{2}{*}{{Method}}&\multirow{2}{*}{{Params(M)}}& \multicolumn{2}{c}{ADE20K} \\ & & & {{GFLOPs $\downarrow$}} & {{mIoU (\%) $\uparrow$}} \\

\midrule

\multirow{2}{*}{ConvNeXt \cite{liu2022convnet}} &UperNet \cite{xiao2018unified} &60.2 &234.7 &46.1 \\
&\textbf{MetaSeg} (Ours) &\textbf{37.2} &\textbf{31.0} &\textbf{46.1} \\

\midrule

\multirow{2}{*}{MobileNetV2 \cite{sandler2018mobilenetv2}} &DeepLabV3 \cite{chen2017rethinking} &18.7 &75.4 &34.1 \\
&\textbf{MetaSeg (Ours)} &\textbf{3.4} &\textbf{4.6} &\textbf{34.7} \\

\bottomrule[1.4pt]
\end{tabular}
}
\caption{Ablation study on the effect of our proposed decoder for other CNN-based backbones on ADE20K validation set.}
\label{table_3}
\end{table}

\begin{table}[t]
\centering
\renewcommand{\arraystretch}{}
\resizebox{\columnwidth}{!}{
\begin{tabular}{ccc|c|cc}
\toprule[1.2pt]

\multirow{2}{*}{{Stage2}}&\multirow{2}{*}{{Stage3}}&\multirow{2}{*}{{Stage4}}&\multirow{2}{*}{{Params(M)}}& \multicolumn{2}{c}{ADE20K} \\ & & & & {{GFLOPs $\downarrow$}} & {{mIoU (\%) $\uparrow$}} \\

\midrule
\cmark &\cmark &\cmark &\textbf{4.7} &\textbf{5.5} &\textbf{42.4}\\
\cmark&\cmark & &4.3 & 5.4 &40.4  \\
&\cmark &\cmark &4.7 & 5.4 &41.6  \\
   \cmark&  & & 4.0 & 5.3 &40.4\\ 
  &  \cmark& & 4.2 & 5.3 & 41.0\\
 & &\cmark & 4.5 & 5.3 & 41.4\\

\bottomrule[1.2pt]
\end{tabular}
}
\caption{Ablation study for applying our proposed Global Meta Block to different stages.}
\label{table_5}
\end{table}

\noindent
\textbf{Speed Benchmark Comparison.} In Table \ref{table_6}, we present the speed benchmark comparisons without any additional accelerating techniques. For fair comparison, we measured Frames Per Second (FPS) of a whole single image of 1536$\times$768 on Cityscapes using a single RTX3090 GPU. Compared to previous methods, our method achieved superior FPS with a higher mIoU score. This result demonstrates that a decrease in FLOPs of our method can lead to improvements in processing speed within the GPU.

\noindent
\textbf{Synapse dataset.}
In Table \ref{table_2}, we compared our MetaSeg with the previous methods on Synapse dataset using DSC (\%). For a fair comparison, we utilized MetaSeg-B in the  medical image segmentation task by considering the similar model size with the previous methods. As shown in Table \ref{table_2}, our MetaSeg-B sets the new state-of-the-art result with 82.78\% DSC. This result showed a 2.09\% higher DSC compared to HiFormer \cite{heidari2023hiformer}. This indicates that our MetaSeg is effective even for the medical image segmentation task. Therefore, we demonstrated the high capabilities of our MetaSeg for application fields.

\subsection{Ablation Study}
\noindent
\textbf{Effectiveness of MetaSeg Decoder for Various CNN-based Backbones.}
In Table \ref{table_3}, we experimented with other CNN-based backbones to evaluate the effect of our MetaSeg decoder. In semantic segmentation, ConvNeXt \cite{liu2022convnet} adopts UperNet \cite{xiao2018unified} as its decoder and MobileNetV2 \cite{sandler2018mobilenetv2} adopts DeepLabV3 \cite{chen2017rethinking} as its decoder. For these CNN-based backbones, our decoder showed competitive performance with significant computational reduction of 86\% and 93.9\%. This indicates that our MetaSeg decoder is an efficient and effective architecture for various CNN-based backbone by enhancing the visual representation from encoder features.  

\begin{table}[t]
\centering
\renewcommand{\arraystretch}{}
\resizebox{\columnwidth}{!}{ \scriptsize
\begin{tabular}{L{2.4cm}|c|cc}
\toprule[0.9pt]

\multirow{2}{*}{{Token Mixer}}&\multirow{2}{*}{{Params (M)}}& \multicolumn{2}{c}{ADE20K} \\ & & {{GFLOPs $\downarrow$}} & {{mIoU (\%) $\uparrow$}} \\

\midrule
AvgPool &4.4 &5.4 &40.7 \\
DW Conv &4.4 &5.4 &40.4 \\
Conv &5.3 &5.8 &41.1 \\
SRA \cite{wang2021pyramid} &5.7 &5.6 &42.4 \\
\midrule
\textbf{CRA (Ours)} &{4.7} & \textbf{5.5} & \textbf{42.4} \\
\bottomrule[0.9pt]
\end{tabular}
} 
\caption{Ablation on the effect of our CRA by applying various token mixers to our Global Meta Block of the decoder. For a fair comparison, we utilized the same backbone, MSCAN-T \cite{guo2022segnext}. }
\label{table_4}
\end{table}

\begin{table}[t]
\centering
\renewcommand{\arraystretch}{}
\resizebox{\columnwidth}{!}{
\begin{tabular}{l|c|c|cc|c}
\toprule[1.2pt]
\multirow{2}{*}{Model} & \multirow{2}{*}{Token Mixer} & \multirow{2}{*}{Params (M)} & \multicolumn{2}{c|}{FLOPs $\downarrow$} & \multirow{2}{*}{mIoU (\%) $\uparrow$}  \\
&&&Attention (M)&Total (G)&\\
\midrule
\multirow{2}{*}{MetaSeg-T} &SRA \cite{wang2021pyramid}  & 5.7 & 62.9 &5.6 &42.4 \\
 &{CRA} (Ours)  & \textbf{4.7} &  \textbf{32.4} \textcolor{hydra_attention_color}{\textbf{(-48.5\%)}}  &\textbf{5.5} &\textbf{42.4}  \\

\midrule
\multirow{2}{*}{MetaSeg-B} &SRA \cite{wang2021pyramid}  & 33.7 & 125.8 &31.1 &48.0  \\
 & {CRA} (Ours)  & \textbf{29.6} & \textbf{63.9} \textcolor{hydra_attention_color}{\textbf{(-49.2\%)}} & \textbf{30.4}&\textbf{48.5}  \\
\bottomrule[1.2pt]
\end{tabular}
} 
\caption{Comparison our CRA with SRA \cite{wang2021pyramid} when applied to MetaSeg-T and MetaSeg-B as a token mixer on ADE20K.} 

\label{table1_rebuttal}
\end{table}

\noindent
\textbf{Effectiveness of Global Meta Block.} In Table \ref{table_5}, we verified the effectiveness of applying GMB in the decoder. We conducted experiments on various cases of applying or non-applying GMB to each Stage\{2,3,4\}. Following \cite{guo2022segnext}, we excluded the features from the first stage of the encoder in this experiment since they contain too much low-level information which degrades the segmentation performance. The results show that applying GMB to Stage\{2,3,4\} is most effective structure compared to other cases. Especially, compared to Stage\{3,4\}, applying GMB to Stage\{2,3,4\} achieved 0.8\% higher mIoU performance even though the parameters and GFLOPs are almost the same. This result indicates that capturing the global contexts through the GMB from all features extracted by the encoder Stage\{2,3,4\} is effective in improving the semantic segmentation performance.



\noindent
\textbf{Effectiveness of Global Modeling Token Mixer in Decoder.}
In Table \ref{table_4}, we conducted an experiment on applying various token mixers to our proposed meta block-based decoder. Through this experiment, we verify which token mixer is the most effective and efficient structure for the decoder when using MSCAN-T, a CNN-based backbone. The global context modeling token mixer (e.g. SRA and our CRA) showed the better mIoU performance compared to the local context modeling token mixer (e.g. pooling, depth-wise convolution and conventional convolution). 
This result demonstrates the importance of considering the global contexts in the decoder when using a CNN-based backbone. 

\begin{figure}[t]
\centering
\includegraphics[width=0.97\linewidth]{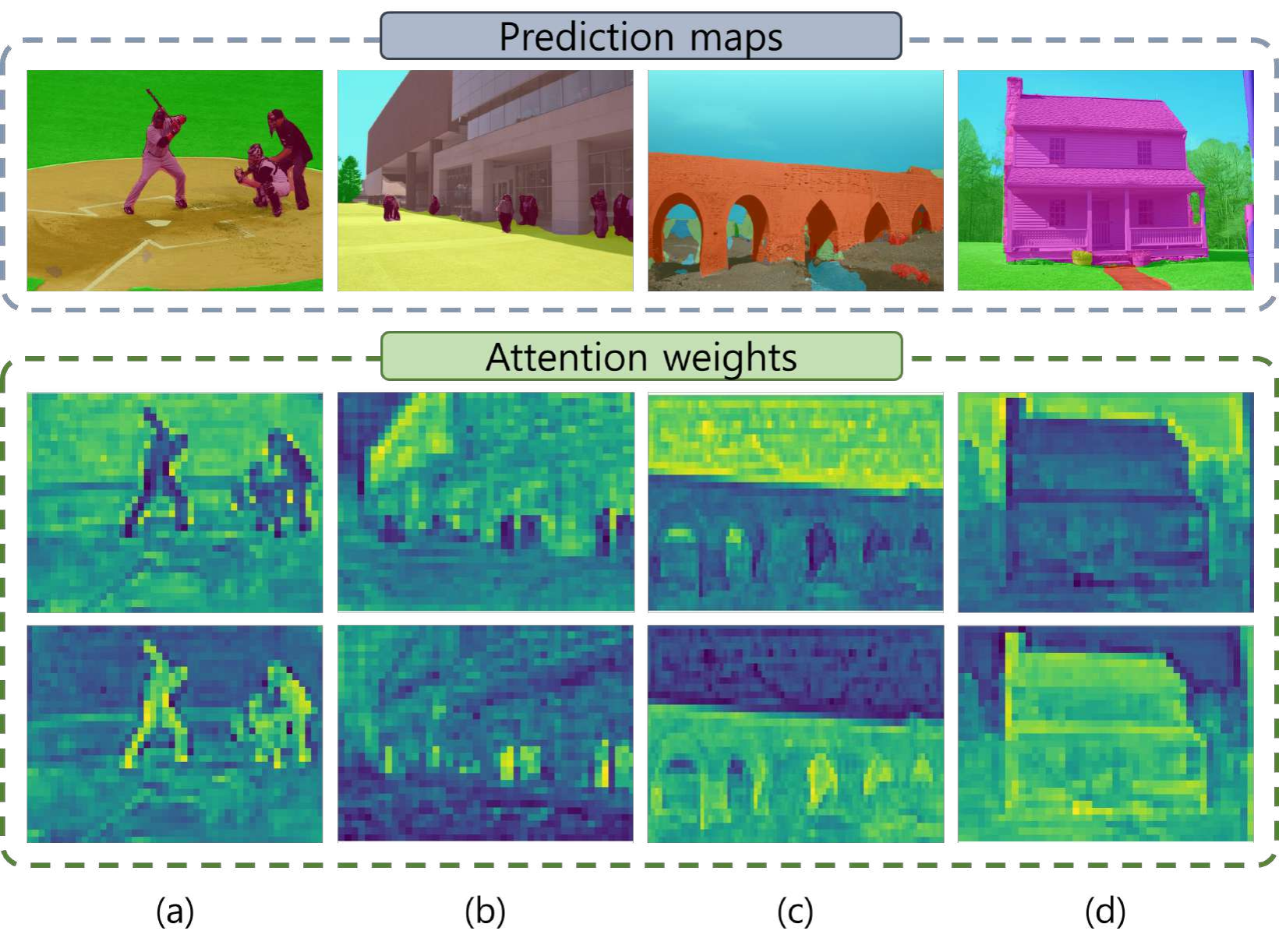}\vspace{-0.3cm}
\caption{Visualization of our prediction maps and our attention score maps on ADE20K.}
\label{attn_vis}
\end{figure}

\begin{figure}[t]
\centering
\includegraphics[width=\linewidth]{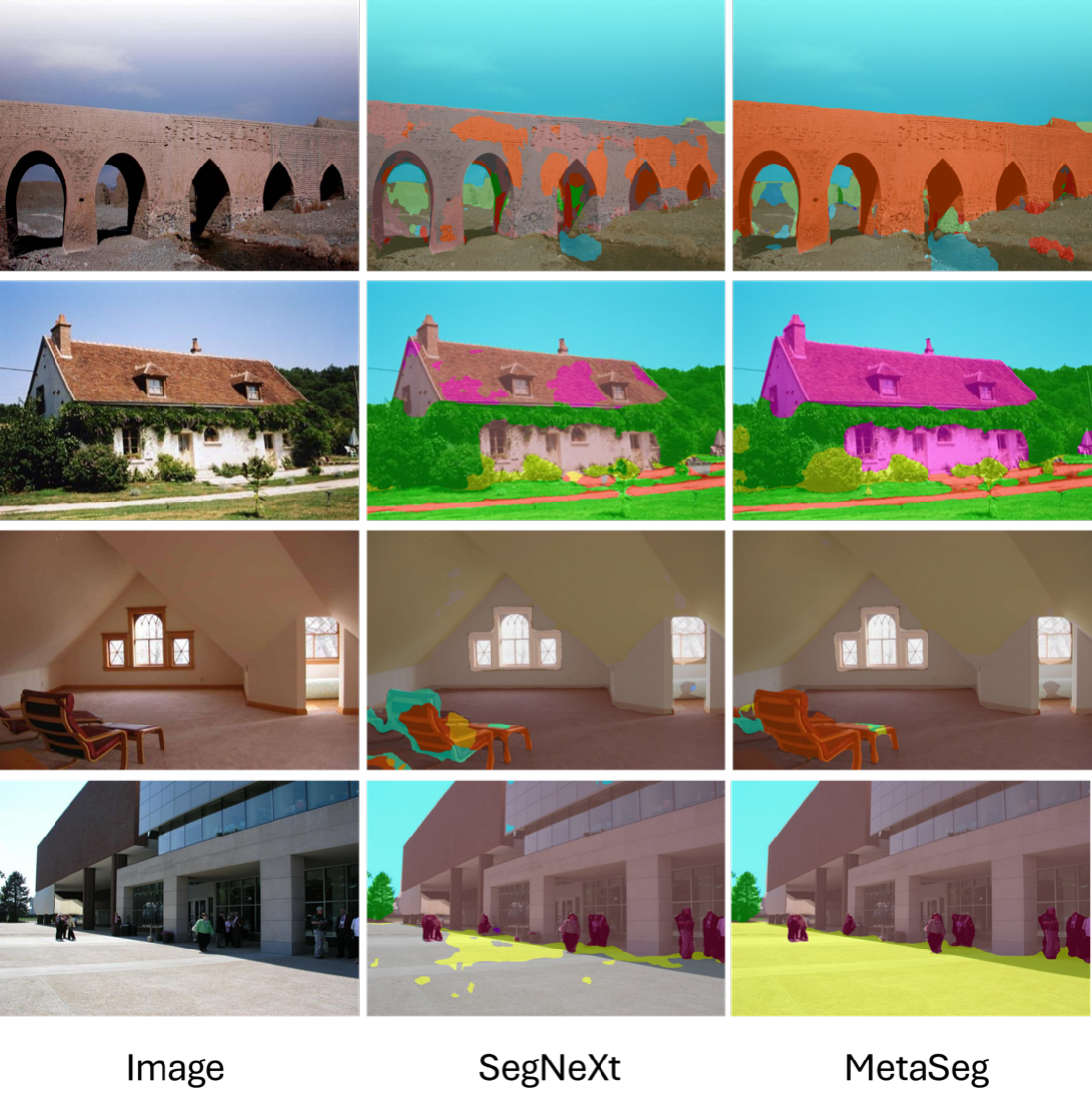} \vspace{-0.5cm}
\caption{Qualitative results on ADE20K dataset. Compared to SegNeXt \cite{guo2022segnext}, our MetaSeg predicts more detailed for various categories.}
\label{qual_ade}
\end{figure}

\noindent
\textbf{Efficiency of Channel Reduction Attention.} In Table \ref{table1_rebuttal}, we focus on the parameter size and computational costs of our channel reduction self-attention (CRA) and the spatial reduction self-attention (SRA) \cite{wang2021pyramid} to compare which method is more efficient in terms of capturing global contexts. SRA is a widely used self-attention method that reduces the spatial resolution of the key-value by treating the token as a vector. In contrast, our CRA scalarizes each query and key token by reducing the channel dimension of the query and key into the one dimension. As shown in Table \ref{table1_rebuttal}, our CRA reduces the computations of the query-key operation by a factor of C times, leading to a total computation reduction for the attention operation that is about twice as much as SRA. For a more detailed comparison of computations as described in eq.(4), we calculated the sum of the computations only for the attention operations in all stages of the decoder. As shown in Table \ref{table1_rebuttal}, the attention operation of our CRA has 48\% and 49\% less FLOPs than the SRA on MetaSeg-T and MetaSeg-B, respectively. 
This indicates that our CRA is more efficient than the previous attention methods, as well as capturing the global context effectively.

\noindent 
\textbf{Visualization of Features.} In Fig. \ref{attn_vis}, we visualized the prediction map and the attention score map of our MetaSeg-T. The attention score map is the similarity score between the query and key, which are applied our channel reduction attention method. As shown in Fig. \ref{attn_vis} (a) and (b), the attention score maps showed significant similarity for people who are far apart. In Fig. \ref{attn_vis} (c) and (d), the similarities of the large regions, such as a bridge and a house, were also captured clearly. These results indicate that our CRA can capture the meaningful similarity scores for extracting the global context features, even though the channel dimension of each pixel token has been reduced to the one dimension. By considering the globality well, our final prediction maps showed accurate segmentation results for the distant objects and the large regions.


\begin{figure}[t]
\centering
\includegraphics[width=\linewidth]{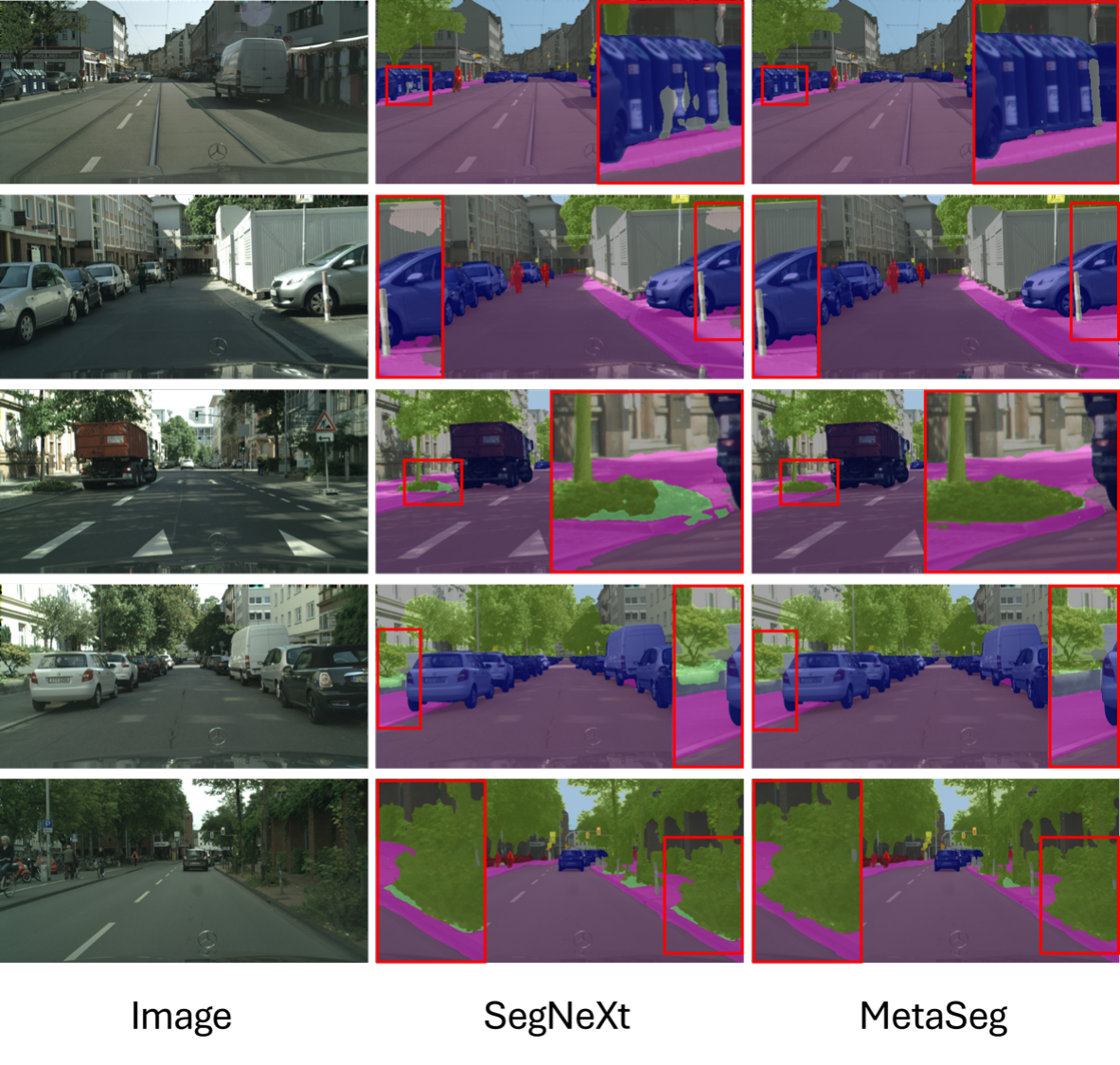}  \vspace{-0.5cm}
\caption{Qualitative results on Cityscapes dataset. The predictions of our MetaSeg are more precise than those of SegNeXt \cite{guo2022segnext}.}
\label{qual_city}
\end{figure}

\begin{figure}[t]
\centering
\includegraphics[width=\linewidth]{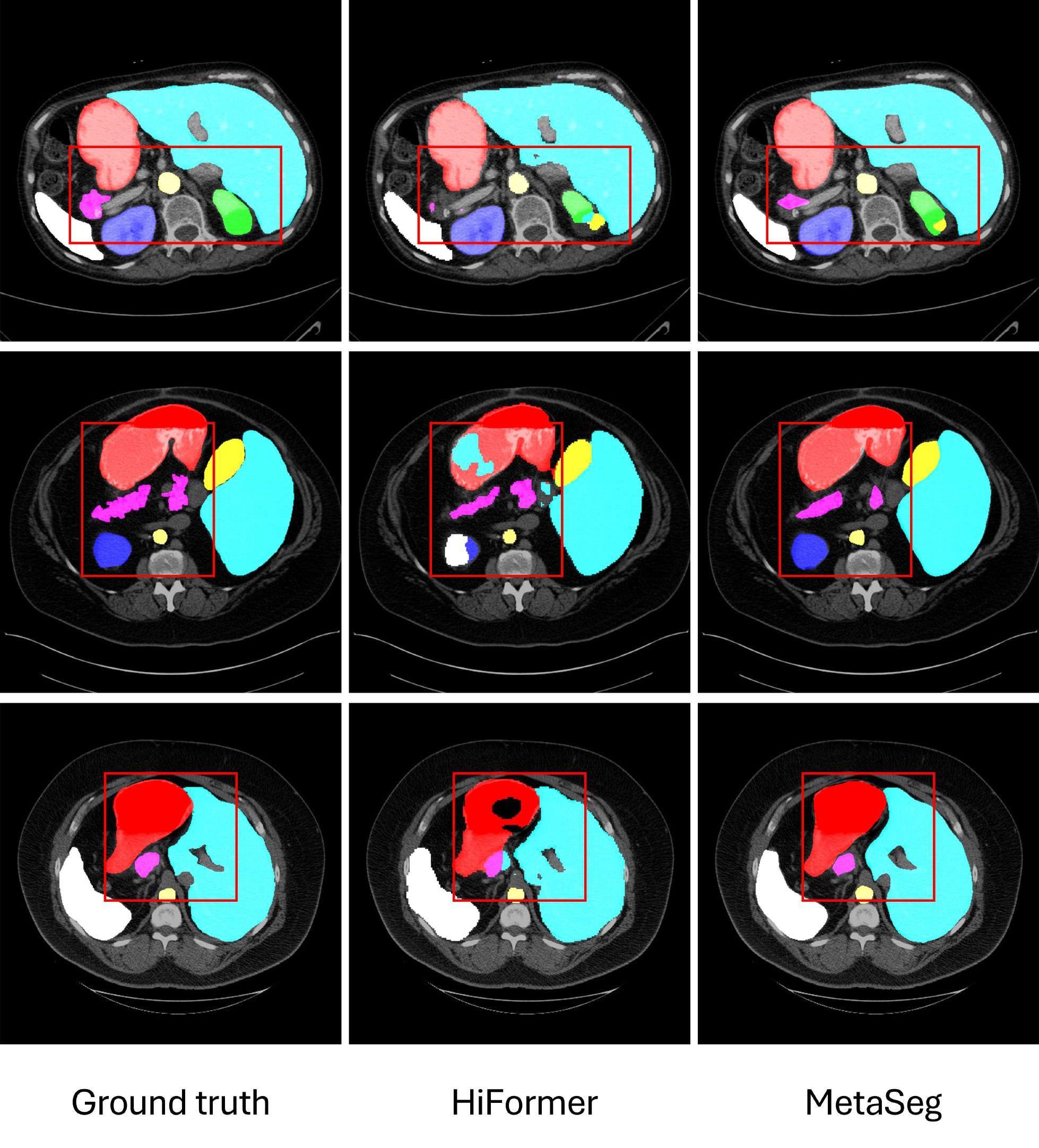} \vspace{-0.5cm}
\caption{Qualitative results on Synapse dataset. Compared to HiFormer \cite{heidari2023hiformer}, the more elaborately predicted regions are highlighted with a red rectangle.}
\label{qual_synapse}
\end{figure}


\subsection{Qualitative Results}
In Figs. \ref{qual_ade} and \ref{qual_city}, we showed segmentation results of our MetaSeg and SegNext \cite{guo2022segnext} on ADE20K and Cityscapes, respectively. Compared to SegNext, our MetaSeg better recognized the object details near the boundaries. This indicates that our model captures more useful visual contexts by leveraging the capacity of the MetaFormer architecture up to the decoder. In addition, our method segmented the large regions (e.g. road and bridge) more precisely. Furthermore, our model showed accurate predictions for far apart objects (e.g. person and house) that belong to the same category. These results indicate that our CRA can sufficiently consider the global contexts with the computational efficiency. In Fig. \ref{qual_synapse}, we compared our predictions with HiFormer \cite{heidari2023hiformer} on Synapse dataset. Our method predicted more accurately for the detailed regions. This indicates that our MetaSeg is effective for the application domain.

\section{Conclusion}
This paper proposed MetaSeg, a novel and powerful semantic segmentation network that effectively captures the local to global contexts by leveraging the MetaFormer architecture up to the decoder. Our MetaSeg showed that the capacity of the MetaFormer can be extended to the decoder as well as the backbone. In addition, we proposed a novel attention module for efficient semantic segmentation, Channel Reduction Attention (CRA) module, which can efficiently consider the globality by reducing the channel dimension of the query and key into the one dimension for low computational costs in the self-attention operation. Experiments demonstrated the effectiveness and efficiency of our method on three public semantic segmentation datasets and a medical image segmentation dataset for application.

\section*{Acknowledgements}
This research was supported by Samsung Electronics Co., Ltd(IO201218-08232-01) and the MSIT(Ministry of Science and ICT), Korea, under the ITRC(Information Technology Research Center) support program(IITP-2023-RS-2023-00260091) supervised by the IITP(Institute for Information \& Communications Technology Planning \& Evaluation).

{\small
\bibliographystyle{ieee_fullname}
\bibliography{egbib}

\begin{thebibliography}{10}\itemsep=-1pt

\bibitem{caesar2018coco}
Holger Caesar, Jasper Uijlings, and Vittorio Ferrari.
\newblock Coco-stuff: Thing and stuff classes in context.
\newblock In {\em Proceedings of the IEEE conference on computer vision and pattern recognition}, pages 1209--1218, 2018.

\bibitem{cao2022swin}
Hu Cao, Yueyue Wang, Joy Chen, Dongsheng Jiang, Xiaopeng Zhang, Qi Tian, and Manning Wang.
\newblock Swin-unet: Unet-like pure transformer for medical image segmentation.
\newblock In {\em European conference on computer vision}, pages 205--218. Springer, 2022.

\bibitem{chen2021transunet}
Jieneng Chen, Yongyi Lu, Qihang Yu, Xiangde Luo, Ehsan Adeli, Yan Wang, Le Lu, Alan~L Yuille, and Yuyin Zhou.
\newblock Transunet: Transformers make strong encoders for medical image segmentation.
\newblock {\em arXiv preprint arXiv:2102.04306}, 2021.

\bibitem{chen2017rethinking}
Liang-Chieh Chen, George Papandreou, Florian Schroff, and Hartwig Adam.
\newblock Rethinking atrous convolution for semantic image segmentation.
\newblock {\em arXiv preprint arXiv:1706.05587}, 2017.

\bibitem{chen2018encoder}
Liang-Chieh Chen, Yukun Zhu, George Papandreou, Florian Schroff, and Hartwig Adam.
\newblock Encoder-decoder with atrous separable convolution for semantic image segmentation.
\newblock In {\em Proceedings of the European conference on computer vision (ECCV)}, pages 801--818, 2018.

\bibitem{cheng2021per}
Bowen Cheng, Alex Schwing, and Alexander Kirillov.
\newblock Per-pixel classification is not all you need for semantic segmentation.
\newblock {\em Advances in Neural Information Processing Systems}, 34:17864--17875, 2021.

\bibitem{cho2022class}
Yubin Cho and Sukju Kang.
\newblock Class attention transfer for semantic segmentation.
\newblock In {\em 2022 IEEE 4th International Conference on Artificial Intelligence Circuits and Systems (AICAS)}, pages 41--45. IEEE, 2022.

\bibitem{cho2023cross}
Yubin Cho, Hyunwoo Yu, and Suk-Ju Kang.
\newblock Cross-aware early fusion with stage-divided vision and language transformer encoders for referring image segmentation.
\newblock {\em IEEE Transactions on Multimedia}, 2023.

\bibitem{cordts2016cityscapes}
Marius Cordts, Mohamed Omran, Sebastian Ramos, Timo Rehfeld, Markus Enzweiler, Rodrigo Benenson, Uwe Franke, Stefan Roth, and Bernt Schiele.
\newblock The cityscapes dataset for semantic urban scene understanding.
\newblock In {\em Proceedings of the IEEE conference on computer vision and pattern recognition}, pages 3213--3223, 2016.

\bibitem{dosovitskiy2020image}
Alexey Dosovitskiy, Lucas Beyer, Alexander Kolesnikov, Dirk Weissenborn, Xiaohua Zhai, Thomas Unterthiner, Mostafa Dehghani, Matthias Minderer, Georg Heigold, Sylvain Gelly, et~al.
\newblock An image is worth 16x16 words: Transformers for image recognition at scale.
\newblock {\em arXiv preprint arXiv:2010.11929}, 2020.

\bibitem{fu2020domain}
Shuhao Fu, Yongyi Lu, Yan Wang, Yuyin Zhou, Wei Shen, Elliot Fishman, and Alan Yuille.
\newblock Domain adaptive relational reasoning for 3d multi-organ segmentation.
\newblock In {\em Medical Image Computing and Computer Assisted Intervention--MICCAI 2020: 23rd International Conference, Lima, Peru, October 4--8, 2020, Proceedings, Part I 23}, pages 656--666. Springer, 2020.

\bibitem{graham2021levit}
Benjamin Graham, Alaaeldin El-Nouby, Hugo Touvron, Pierre Stock, Armand Joulin, Herv{\'e} J{\'e}gou, and Matthijs Douze.
\newblock Levit: a vision transformer in convnet's clothing for faster inference.
\newblock In {\em Proceedings of the IEEE/CVF international conference on computer vision}, pages 12259--12269, 2021.

\bibitem{guo2022segnext}
Meng-Hao Guo, Cheng-Ze Lu, Qibin Hou, Zhengning Liu, Ming-Ming Cheng, and Shi-Min Hu.
\newblock Segnext: Rethinking convolutional attention design for semantic segmentation.
\newblock {\em arXiv preprint arXiv:2209.08575}, 2022.

\bibitem{heidari2023hiformer}
Moein Heidari, Amirhossein Kazerouni, Milad Soltany, Reza Azad, Ehsan~Khodapanah Aghdam, Julien Cohen-Adad, and Dorit Merhof.
\newblock Hiformer: Hierarchical multi-scale representations using transformers for medical image segmentation.
\newblock In {\em Proceedings of the IEEE/CVF Winter Conference on Applications of Computer Vision}, pages 6202--6212, 2023.

\bibitem{landman2015synapse}
Bennett Landman, Zhoubing Xu, Juan Igelsias, Martin Styner, T Langerak, and Arno Klein.
\newblock Miccai multi-atlas labeling beyond the cranial vault-workshop and challenge.
\newblock In {\em MICCAI}, 2015.

\bibitem{li2022efficientformer}
Yanyu Li, Geng Yuan, Yang Wen, Ju Hu, Georgios Evangelidis, Sergey Tulyakov, Yanzhi Wang, and Jian Ren.
\newblock Efficientformer: Vision transformers at mobilenet speed.
\newblock {\em Advances in Neural Information Processing Systems}, 35:12934--12949, 2022.

\bibitem{liu2021swin}
Ze Liu, Yutong Lin, Yue Cao, Han Hu, Yixuan Wei, Zheng Zhang, Stephen Lin, and Baining Guo.
\newblock Swin transformer: Hierarchical vision transformer using shifted windows.
\newblock In {\em Proceedings of the IEEE/CVF international conference on computer vision}, pages 10012--10022, 2021.

\bibitem{liu2022convnet}
Zhuang Liu, Hanzi Mao, Chao-Yuan Wu, Christoph Feichtenhofer, Trevor Darrell, and Saining Xie.
\newblock A convnet for the 2020s.
\newblock In {\em Proceedings of the IEEE/CVF Conference on Computer Vision and Pattern Recognition}, pages 11976--11986, 2022.

\bibitem{long2015fully}
Jonathan Long, Evan Shelhamer, and Trevor Darrell.
\newblock Fully convolutional networks for semantic segmentation.
\newblock In {\em Proceedings of the IEEE conference on computer vision and pattern recognition}, pages 3431--3440, 2015.

\bibitem{milletari2016v}
Fausto Milletari, Nassir Navab, and Seyed-Ahmad Ahmadi.
\newblock V-net: Fully convolutional neural networks for volumetric medical image segmentation.
\newblock In {\em 2016 fourth international conference on 3D vision (3DV)}, pages 565--571. Ieee, 2016.

\bibitem{ronneberger2015u}
Olaf Ronneberger, Philipp Fischer, and Thomas Brox.
\newblock U-net: Convolutional networks for biomedical image segmentation.
\newblock In {\em Medical Image Computing and Computer-Assisted Intervention--MICCAI 2015: 18th International Conference, Munich, Germany, October 5-9, 2015, Proceedings, Part III 18}, pages 234--241. Springer, 2015.

\bibitem{sandler2018mobilenetv2}
Mark Sandler, Andrew Howard, Menglong Zhu, Andrey Zhmoginov, and Liang-Chieh Chen.
\newblock Mobilenetv2: Inverted residuals and linear bottlenecks.
\newblock In {\em Proceedings of the IEEE conference on computer vision and pattern recognition}, pages 4510--4520, 2018.

\bibitem{schlemper2019attention}
Jo Schlemper, Ozan Oktay, Michiel Schaap, Mattias Heinrich, Bernhard Kainz, Ben Glocker, and Daniel Rueckert.
\newblock Attention gated networks: Learning to leverage salient regions in medical images.
\newblock {\em Medical image analysis}, 53:197--207, 2019.

\bibitem{shim2023feedformer}
Jae-hun Shim, Hyunwoo Yu, Kyeongbo Kong, and Suk-ju Kang.
\newblock Feedformer: Revisiting transformer decoder for efficient semantic segmentation.
\newblock {\em Proceedings of the AAAI Conference on Artificial Intelligence}, pages 2263--2271, 2023.

\bibitem{tolstikhin2021mlp}
Ilya~O Tolstikhin, Neil Houlsby, Alexander Kolesnikov, Lucas Beyer, Xiaohua Zhai, Thomas Unterthiner, Jessica Yung, Andreas Steiner, Daniel Keysers, Jakob Uszkoreit, et~al.
\newblock Mlp-mixer: An all-mlp architecture for vision.
\newblock {\em Advances in neural information processing systems}, 34:24261--24272, 2021.

\bibitem{touvron2022resmlp}
Hugo Touvron, Piotr Bojanowski, Mathilde Caron, Matthieu Cord, Alaaeldin El-Nouby, Edouard Grave, Gautier Izacard, Armand Joulin, Gabriel Synnaeve, Jakob Verbeek, et~al.
\newblock Resmlp: Feedforward networks for image classification with data-efficient training.
\newblock {\em IEEE Transactions on Pattern Analysis and Machine Intelligence}, 2022.

\bibitem{vaswani2017attention}
Ashish Vaswani, Noam Shazeer, Niki Parmar, Jakob Uszkoreit, Llion Jones, Aidan~N Gomez, {\L}ukasz Kaiser, and Illia Polosukhin.
\newblock Attention is all you need.
\newblock {\em Advances in neural information processing systems}, 30, 2017.

\bibitem{wang2022mixed}
Hongyi Wang, Shiao Xie, Lanfen Lin, Yutaro Iwamoto, Xian-Hua Han, Yen-Wei Chen, and Ruofeng Tong.
\newblock Mixed transformer u-net for medical image segmentation.
\newblock In {\em ICASSP 2022-2022 IEEE International Conference on Acoustics, Speech and Signal Processing (ICASSP)}, pages 2390--2394. IEEE, 2022.

\bibitem{wang2023internimage}
Wenhai Wang, Jifeng Dai, Zhe Chen, Zhenhang Huang, Zhiqi Li, Xizhou Zhu, Xiaowei Hu, Tong Lu, Lewei Lu, Hongsheng Li, et~al.
\newblock Internimage: Exploring large-scale vision foundation models with deformable convolutions.
\newblock In {\em Proceedings of the IEEE/CVF Conference on Computer Vision and Pattern Recognition}, pages 14408--14419, 2023.

\bibitem{wang2021pyramid}
Wenhai Wang, Enze Xie, Xiang Li, Deng-Ping Fan, Kaitao Song, Ding Liang, Tong Lu, Ping Luo, and Ling Shao.
\newblock Pyramid vision transformer: A versatile backbone for dense prediction without convolutions.
\newblock In {\em Proceedings of the IEEE/CVF international conference on computer vision}, pages 568--578, 2021.

\bibitem{wang2022pvt}
Wenhai Wang, Enze Xie, Xiang Li, Deng-Ping Fan, Kaitao Song, Ding Liang, Tong Lu, Ping Luo, and Ling Shao.
\newblock Pvt v2: Improved baselines with pyramid vision transformer.
\newblock {\em Computational Visual Media}, 8(3):415--424, 2022.

\bibitem{wu2021cvt}
Haiping Wu, Bin Xiao, Noel Codella, Mengchen Liu, Xiyang Dai, Lu Yuan, and Lei Zhang.
\newblock Cvt: Introducing convolutions to vision transformers.
\newblock In {\em Proceedings of the IEEE/CVF International Conference on Computer Vision}, pages 22--31, 2021.

\bibitem{xiao2018unified}
Tete Xiao, Yingcheng Liu, Bolei Zhou, Yuning Jiang, and Jian Sun.
\newblock Unified perceptual parsing for scene understanding.
\newblock In {\em Proceedings of the European conference on computer vision (ECCV)}, pages 418--434, 2018.

\bibitem{xie2021segformer}
Enze Xie, Wenhai Wang, Zhiding Yu, Anima Anandkumar, Jose~M Alvarez, and Ping Luo.
\newblock Segformer: Simple and efficient design for semantic segmentation with transformers.
\newblock {\em Advances in Neural Information Processing Systems}, 34:12077--12090, 2021.

\bibitem{yang2022lite}
Chenglin Yang, Yilin Wang, Jianming Zhang, He Zhang, Zijun Wei, Zhe Lin, and Alan Yuille.
\newblock Lite vision transformer with enhanced self-attention.
\newblock In {\em Proceedings of the IEEE/CVF Conference on Computer Vision and Pattern Recognition}, pages 11998--12008, 2022.

\bibitem{yu2024embedding}
Hyunwoo Yu, Yubin Cho, Beoungwoo Kang, Seunghun Moon, Kyeongbo Kong, and Suk-Ju Kang.
\newblock Embedding-free transformer with inference spatial reduction for efficient semantic segmentation.
\newblock {\em arXiv preprint arXiv:2407.17261}, 2024.

\bibitem{yu2022vision}
Hyunwoo Yu, Jae-hun Shim, Jaeho Kwak, Jou~Won Song, and Suk-Ju Kang.
\newblock Vision transformer-based retina vessel segmentation with deep adaptive gamma correction.
\newblock In {\em ICASSP 2022-2022 IEEE International Conference on Acoustics, Speech and Signal Processing (ICASSP)}, pages 1456--1460. IEEE, 2022.

\bibitem{yu2022metaformer}
Weihao Yu, Mi Luo, Pan Zhou, Chenyang Si, Yichen Zhou, Xinchao Wang, Jiashi Feng, and Shuicheng Yan.
\newblock Metaformer is actually what you need for vision.
\newblock In {\em Proceedings of the IEEE/CVF conference on computer vision and pattern recognition}, pages 10819--10829, 2022.

\bibitem{zhang2022topformer}
Wenqiang Zhang, Zilong Huang, Guozhong Luo, Tao Chen, Xinggang Wang, Wenyu Liu, Gang Yu, and Chunhua Shen.
\newblock Topformer: Token pyramid transformer for mobile semantic segmentation.
\newblock In {\em Proceedings of the IEEE/CVF Conference on Computer Vision and Pattern Recognition}, pages 12083--12093, 2022.

\bibitem{zhao2017pyramid}
Hengshuang Zhao, Jianping Shi, Xiaojuan Qi, Xiaogang Wang, and Jiaya Jia.
\newblock Pyramid scene parsing network.
\newblock In {\em Proceedings of the IEEE conference on computer vision and pattern recognition}, pages 2881--2890, 2017.

\bibitem{zheng2021rethinking}
Sixiao Zheng, Jiachen Lu, Hengshuang Zhao, Xiatian Zhu, Zekun Luo, Yabiao Wang, Yanwei Fu, Jianfeng Feng, Tao Xiang, Philip~HS Torr, et~al.
\newblock Rethinking semantic segmentation from a sequence-to-sequence perspective with transformers.
\newblock In {\em Proceedings of the IEEE/CVF conference on computer vision and pattern recognition}, pages 6881--6890, 2021.

\bibitem{zhou2017scene}
Bolei Zhou, Hang Zhao, Xavier Puig, Sanja Fidler, Adela Barriuso, and Antonio Torralba.
\newblock Scene parsing through ade20k dataset.
\newblock In {\em Proceedings of the IEEE conference on computer vision and pattern recognition}, pages 633--641, 2017.

\end{thebibliography}
}

\newpage

\appendix

\section*{Appendix}
\begin{itemize}
    \item In \Cref{sec:A}, we provide additional experiments on adopting Transformer-based encoder as a backbone in the proposed MetaSeg network. 
    \item In \Cref{sec:B}, we provide a comparison of our MetaSeg with other segmentation networks.
    \item In \Cref{sec:C}, we provide additional results on the inference speed (FPS).
    \item In \Cref{sec:D}, we provide additional qualitative results compared with the proposed and previous model on ADE20K, Cityscapes, COCO-Stuff and Synapse datasets.
\end{itemize}

\section{Effectiveness of Our MetaSeg for Various Transformer-based Backbone}
\label{sec:A}
In Table \ref{table_A}, we conducted the experiment on using the Transformer-based encoder as a backbone of our MetaSeg. Previously, Mix Transformer (MiT) \cite{xie2021segformer} and Lite Vision Transformer (LVT) \cite{yang2022lite} backbones adopt SegFormer \cite{xie2021segformer} as its semantic segmentation decoder. Compared to SegFormer \cite{xie2021segformer}, our MetaSeg remarkably reduces the computational costs (GFLOPs) by 53.6 \% and 43.4 \% with mIoU improvements of 1.9\% and 0.4\% for MiT \cite{xie2021segformer} and LVT \cite{yang2022lite}, respectively. These results indicate that the Transformer-based backbones as well as the CNN-based backbones can effectively leverage our MetaSeg for efficient semantic segmentation task.

\section{Comparison of Our MetaSeg with Other Semantic Segmentation Networks}
\label{sec:B}
In Table \ref{tab:tableB}, we compared our method with other segmentation networks to demonstrate the power of our MetaSeg. We experimented with the same backbone for a fair comparison. Compared to other networks, our model showed significant computational reduction with higher mIoU performance. This result indicates that our MetaSeg is a powerful and efficient segmentation network by leveraging the MetaFormer block that uses our efficient CRA module as a token mixer. 

\section{Inference Speed Comparison}
\label{sec:C}
In Table \ref{table_C}, we represent the inference speed comparisons under the mmsegmentation code base without any additional accelerating techniques. We tested Frame Per Second (FPS) of a single image of 1024 $\times$ 2048 on Cityscapes test dataset using a single RTX3090 GPU. The results show that our MetaSeg is fastest compared with other lightweight semantic segmentation models \cite{guo2022segnext,shim2023feedformer,xie2021segformer}, while achieving the highest mIoU performance.

\begin{table}[t]
\centering
\renewcommand{\arraystretch}{1.1}
\resizebox{\columnwidth}{!}{
\begin{tabular}{l|c|c|cc}
\toprule[1.4pt]

\multirow{2}{*}{{Backbone}}&\multirow{2}{*}{{Method}}&\multirow{2}{*}{{Params (M)}}& \multicolumn{2}{c}{ADE20K} \\ & & & {{GFLOPs $\downarrow$}} & {{mIoU (\%) $\uparrow$}} \\

\midrule

\multirow{2}{*}{MiT \cite{xie2021segformer}} &SegFormer \cite{xie2021segformer} &3.8 &8.4 &37.4 \\

&\textbf{MetaSeg (Ours)} &\textbf{4.1} &\textbf{3.9} &\textbf{39.3} \\

\midrule

\multirow{2}{*}{LVT \cite{yang2022lite}} &SegFormer \cite{xie2021segformer} &3.9 &10.6 &39.3 \\
&\textbf{MetaSeg (Ours)} &\textbf{4.2} &\textbf{6.0} &\textbf{39.7} \\

\bottomrule[1.4pt]
\end{tabular}
}

\caption{Effectiveness of our MetaSeg for Transformer-based backbones on ADE20K validation set.}
\label{table_A}
\end{table}

\begin{table}[t]
\centering
\renewcommand{\arraystretch}{}
\resizebox{\columnwidth}{!}{%
\begin{tabular}{l|c|c|c}
\toprule[1.4pt]
{Method}  &  Params (M) & GFLOPs $\downarrow$ & mIoU (\%) $\uparrow$ \\
\midrule
FCN \cite{long2015fully}& 9.8&39.6&19.7\\
PSPNet \cite{zhao2017pyramid} & 13.7 & 53.0&29.7\\
DeepLabV3 \cite{chen2017rethinking}  & 18.7& 75.4& 34.1\\
DeepLabV3+ \cite{chen2018encoder} & 15.4&69.5& 34.0\\
\midrule
MetaSeg (Ours)  & \textbf{3.4} & \textbf{4.6}& \textbf{34.7}\\
\bottomrule[1.4pt]
\end{tabular}
}
\caption{Comparison of our MetaSeg with other segmentation Networks on ADE20K dataset. For a fair comparison, we use the same backbone, MobileNetV2 \cite{sandler2018mobilenetv2}.}
\label{tab:tableB}
\end{table}

\begin{table}[t]
\centering
\renewcommand{\arraystretch}{1.1}
\resizebox{\columnwidth}{!}{

\begin{tabular}{L{3.5cm}|c|ccc}
\toprule[1.5pt]

\multirow{2}{*}{{Method}}&\multirow{2}{*}{{Params(M)}}& \multicolumn{3}{c}{Cityscapes}\\ & & {{GFLOPs $\downarrow$}} & {{mIoU (\%) $\uparrow$}} & {{FPS (img/s)$\uparrow$}}  \\

\midrule
SegFormer-B0 \cite{xie2021segformer} &3.8 &125.5 &76.2 & 12.52\\
FeedFormer-B0 \cite{shim2023feedformer} &4.5 &107.4 &77.9 & 17.33 \\
SegNeXt-T \cite{guo2022segnext}   &4.3 &50.5 &79.8 & 22.73\\
\midrule
\textbf{MetaSeg-T (Ours)}  &{4.7} &\textbf{47.9} &\textbf{80.1} & \textbf{23.46}\\

\bottomrule[1.5pt]
\end{tabular}
}
\caption{Comparison of our MetaSeg with previous state-of-the-art methods on Cityscapes. }
\label{table_C}
\end{table}

\begin{figure}[t]
\centering
\includegraphics[width=\linewidth]{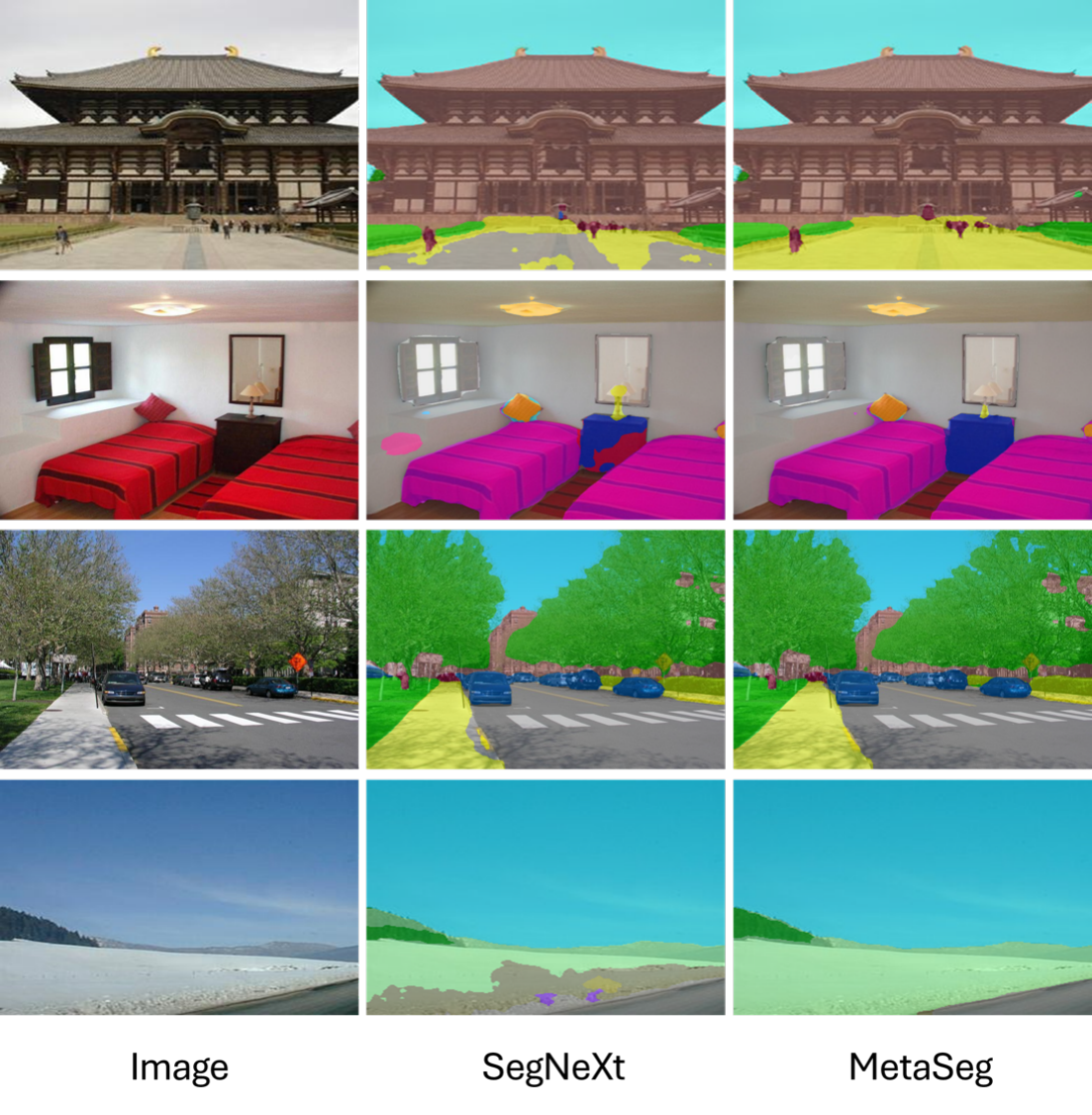}
\caption{Qualitative results on ADE20K. Compared to the previous state-or-the-art method, our MetaSeg generates more accurate segmentation maps across various categories.}

\label{qual_ade_supple}
\end{figure}

\begin{figure}[t] 
\centering
\includegraphics[width=\linewidth]{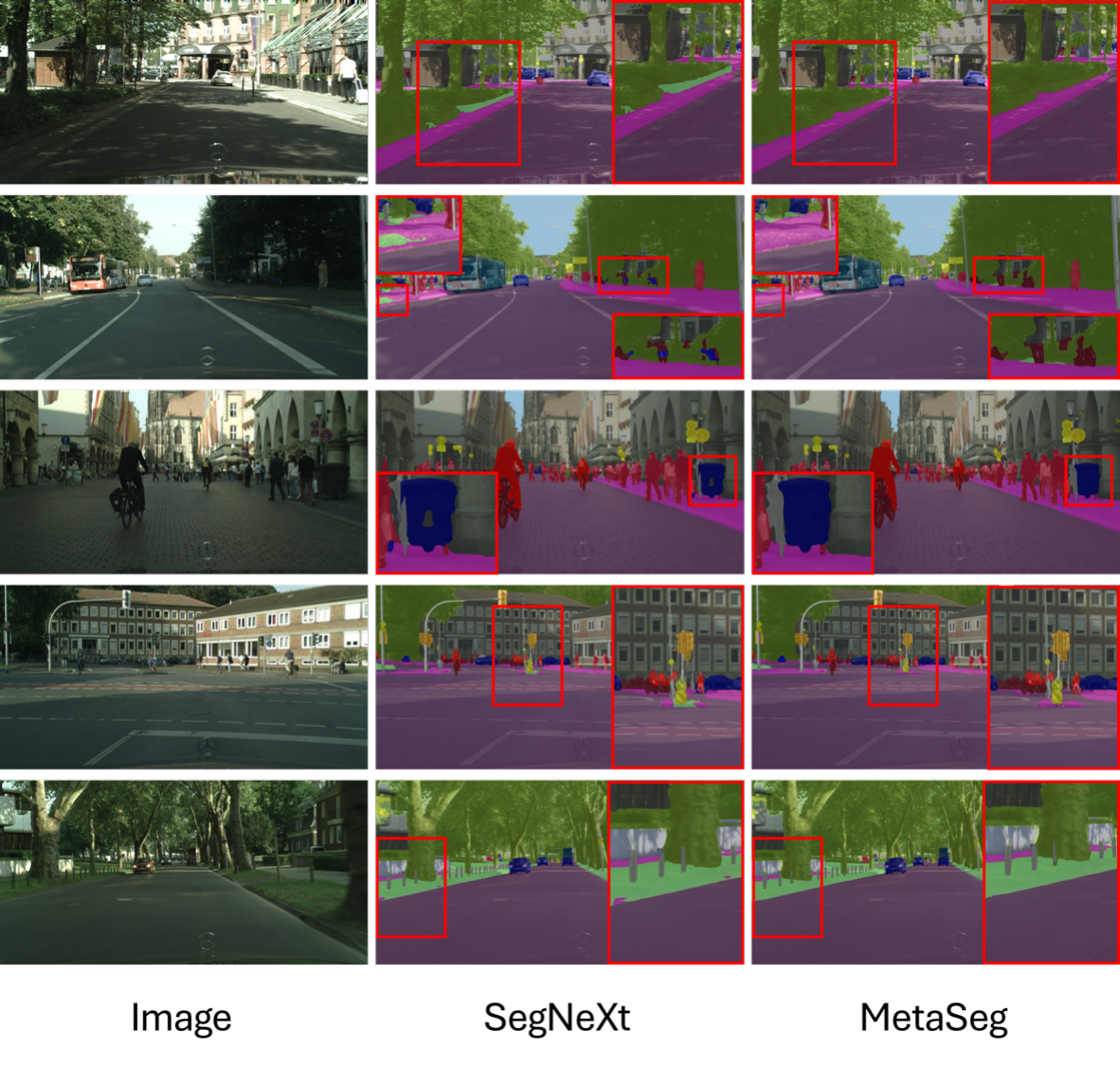}
\caption{Qualitative results on Cityscapes. For multiple categories, our MetaSeg provides more precise predictions than SegNeXt \cite{guo2022segnext}.}

\label{qual_city_supple}
\end{figure}

\begin{figure}[t]
\centering
\includegraphics[width=\linewidth]{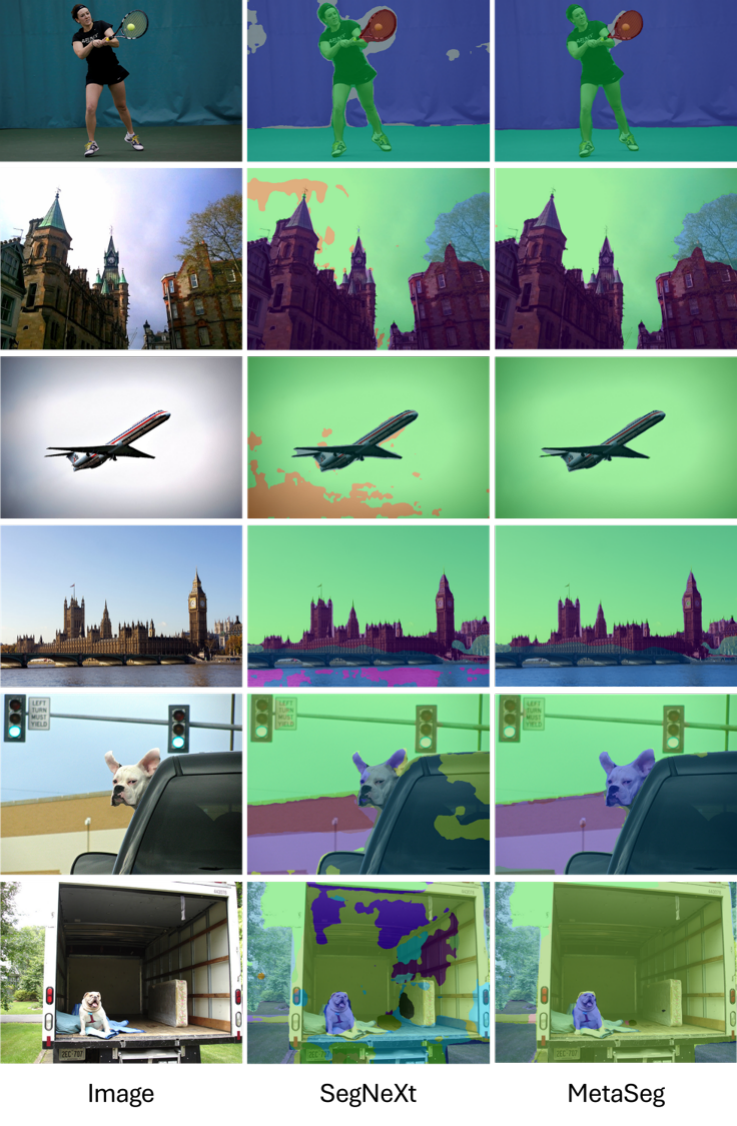}
\caption{Qualitative results on COCO-Stuff. Compared to SegNeXt \cite{guo2022segnext}, our MetaSeg provides more detailed segmentation predictions.}

\label{qual_coco_supple}
\end{figure}

\section{Additional Qualitative Results}
\label{sec:D}
\begin{figure}[t]
\centering
\includegraphics[width=\linewidth]{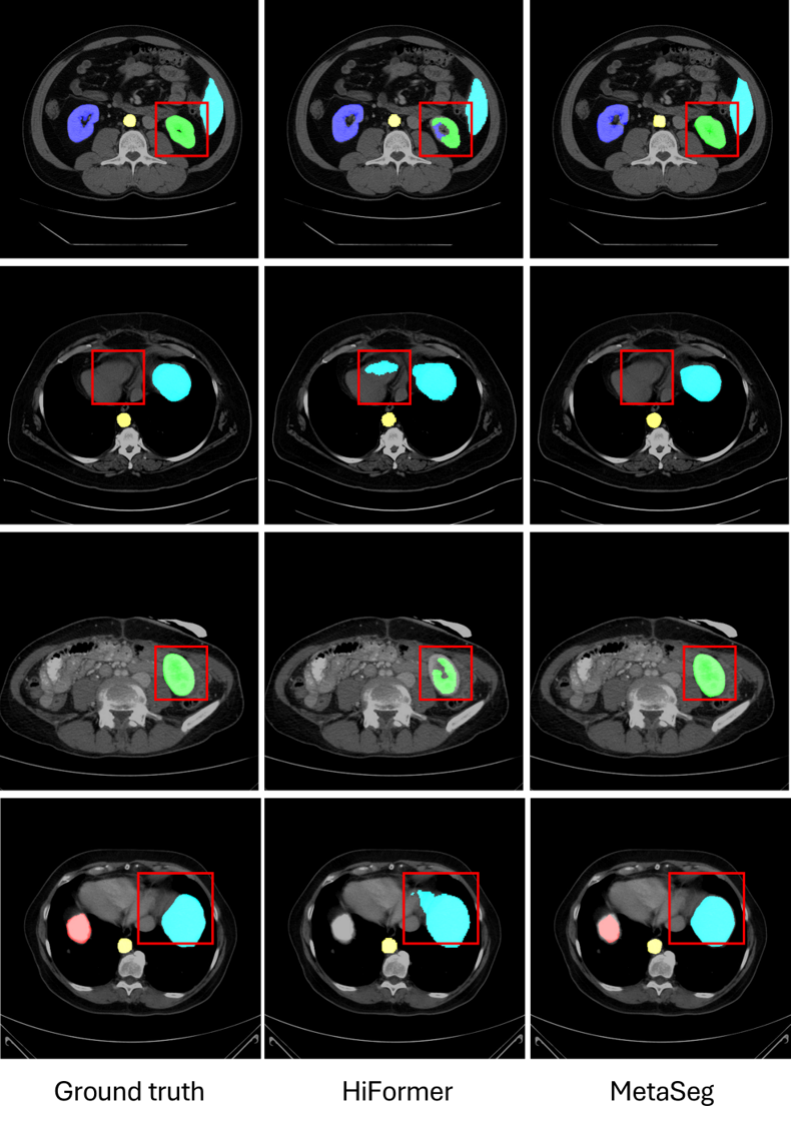}
\caption{Qualitative results on Synapse. Our MetaSeg predicts more precisely than HiFormer \cite{heidari2023hiformer} across various categories.}

\label{qual_synapse_supple}
\end{figure}

In Fig. \ref{qual_ade_supple}, \ref{qual_city_supple} and \ref{qual_coco_supple}, we visualized additional qualitative results of our MetaSeg and the previous state-of-the-art method on ADE20K, Cityscapes, COCO-Stuff datasets, respectively. Compared to SegNeXt \cite{guo2022segnext}, our MetaSeg showed more accurate predictions for large regions. Our MetaSeg also predicted more detailed for the object boundaries than SegNeXt \cite{guo2022segnext}. In addition, we visualized more qualitative results of our MetaSeg and HiFormer \cite{heidari2023hiformer} on Synapse dataset. As shown in Fig. \ref{qual_synapse_supple}, our MetaSeg predicted more precisely than HiFormer \cite{heidari2023hiformer} for various categories. These results indicate that our MetaSeg can effectively capture the local to global information by extensively leveraging the MetaFormer architecture from the encoder to the decoder.

\end{document}